\documentclass[times,twocolumn,final,longtitle]{elsarticle}

\usepackage{jasr}
\usepackage{framed,multirow}

\usepackage{amssymb}
\usepackage{latexsym}

\usepackage[switch]{lineno}

\usepackage{url}
\usepackage{xcolor}
\definecolor{newcolor}{rgb}{.8,.349,.1}

\usepackage[citebordercolor=white]{hyperref}

\usepackage{amsmath}

\makeatletter
\providecommand{\stmtitleheader}{}
\makeatother

\begin{document}

\verso{Morgan Coe \textit{et al.}}

\begin{frontmatter}

\title{Persistent feature reconstruction of resident space objects (RSOs) within inverse synthetic aperture radar (ISAR) images}%

\author[1]{Morgan \snm{Coe}\fnref{fn1}}
\author[1]{Gruffudd \snm{Jones}\fnref{fn2}}

\fntext[fn1]{Email: m.coe@pgr.bham.ac.uk}
\fntext[fn2]{Email: g.jones.3@pgr.bham.ac.uk}

\author[1]{Leah-Nani \snm{Alconcel} \corref{cor1}}
\cortext[cor1]{Corresponding author:}
\ead{l.alconcel@bham.ac.uk}

\author[1]{Marina \snm{Gashinova}\fnref{fn3}}
\fntext[fn3]{Email: m.s.gashinova@bham.ac.uk}

\affiliation[1]{organization={University of Birmingham},
                addressline={Edgbaston},
                city={Birmingham},
                postcode={B15 2TT},
                country={United Kingdom}}
\begin{abstract}
With the rapidly growing population of resident space objects (RSOs) in the near-Earth space environment, detailed information about their condition and capabilities is needed to provide Space Domain Awareness (SDA). Space-based sensing will enable inspection of RSOs at shorter ranges, independent of atmospheric effects, and from all aspects. The use of a sub-THz inverse synthetic aperture radar (ISAR) imaging and sensing system for SDA has been proposed in previous work, demonstrating the achievement of sub-cm image resolution at ranges of up to 100 km. This work focuses on recognition of external structures by use of sequential feature detection and tracking throughout the aligned ISAR images of the satellites. The Hough transform is employed to detect linear features, which are tracked throughout the sequence. ISAR imagery is generated via a metaheuristic simulator capable of modelling encounters for a variety of deployment scenarios. Initial frame-to-frame alignment is achieved through a series of affine transformations to facilitate later association between image features. A gradient-by-ratio method is used for edge detection within individual ISAR images, and edge magnitude and direction are subsequently used to inform a double-weighted Hough transform to detect features with high accuracy.  
Feature evolution during sequences of frames is analysed. It is shown that the use of feature tracking within sequences with the proposed approach will increase confidence in feature detection and classification, and an example use-case of robust detection of shadowing as a feature is presented.
\end{abstract}


\end{frontmatter}

\section{Introduction}
Space Domain Awareness (SDA) has become increasingly important as the Earth-orbiting satellite population continues to grow \citep{european_space_agency_esa_2025}. Tracking and surveillance of resident space objects (RSOs) is largely achieved through ground-based radar and optical systems, both of which have limitations in resolution and coverage. The development of novel space-based monitoring systems is crucial to provide SDA and enable the near-Earth space environment to be maintained sustainably.

Ground-based radar such as the Tracking and Imaging Radar System (TIRA) \citep{klare_future_2024} and Haystack Ultra-Wideband Satellite Imaging Radar (HUSIR) \citep{macdonald_husir_2014} can provide a wide field of view, tracking, life pattern estimation, and ISAR imagery. However, for full accurate characterisation, independent of angle and unaffected by atmospheric conditions, imagery from space is needed. ISAR imagery can be performed on cooperative and non-cooperative targets, enabling full characterisation based on semantic and spatial-temporal data.
Use of sub-THz frequencies for space-based ISAR imaging, as demonstrated previously \citep{marchetti_space-based_2022}, permits surfaces that would be electrically smooth at lower frequencies to be resolved, and requires lower operational power compared to ground-based. Additionally, use of radar enables precise determination of RSO distances and velocities at long ranges \citep{chen_inverse_2014}.

Various techniques have been demonstrated to extract information from ISAR imagery. \citet{xue_sequential_2022} use a hybrid transformer to identify a satellite from a sequence of ISAR images. \citet{du_swin_2024} use Swin transformers to apply well-trained optical recognition models to specific ISAR scenarios, with an ISAR simulation method that degrades optical imagery to mimic the sparse returns within ISAR imagery. \citet{li_polarimetric_2025} and \citet{xin_scale-shift_2025} use polarimetric data as a classification tool. 
These methods use supervised learning models pre-trained on large data sets to recognise and classify satellites as whole structures. By contrast, the methodology in this paper uses unsupervised feature detection, requiring no large training data set, and aims to characterise satellites via detection and recognition of individual components and deployables. At the chosen sub-THz frequencies, small features and surface texture are likely to be resolved.




This paper presents an approach for exploiting a sequence of ISAR imagery to obtain information about an RSO's structure and condition for SDA purposes with high confidence. Detection and retention of features throughout a sequence is demonstrated, as well as rejection of false-positive feature detections, and reduction of image artefacts. This process is an essential facilitator for characterisation and classification, without reliance on large data sets and supervised machine learning approaches. 

In Section \ref{sec:setup}, a methodology for obtaining sequences of ISAR imagery is presented, starting with the design of orbital deployments, image simulation corresponding to these deployments, and sequence alignment techniques. In Section \ref{sec:feature-detection}, feature detection via a double-weighted Hough transform is detailed, along with ISAR-specific image preprocessing to enhance feature detection accuracy.
In Section \ref{sec:persistent}, detected features are clustered using the DBSCAN algorithm, which aids tracking of features throughout the sequence.
Classification of shadows through cluster analysis is shown. 
And the utility of cumulative sequence images is discussed.
\section{Methodology}
\label{sec:setup}
\subsection{ISAR system deployment}
\label{sec:encounter-details}
In previous work \citep{jones_strategies_2025}, scenarios for space-based observation of RSOs using sub-THz ISAR imaging from a space-based platform were designed and presented. While that work focuses on a deployment for monitoring RSOs in GEO, the imaging process is also applicable to other orbital regimes without loss of generality \citep{marchetti_space-based_2022}.
Key design elements of the space-borne ISAR system retained from this previous work include a tracking antenna with 30 cm physical aperture diameter operating at centre frequencies of either 75 or 300 GHz, and encounters between the satellite bearing the ISAR sensing system and the target being constrained to a maximum separation of 100 km. 

ISAR imaging is based on the change of the target's aspect angle, and the size of the imaging aperture is defined by the desired image resolution. Therefore, the total change of aspect angle within an encounter (which may be up to 100 seconds) can be subdivided to produce a sequence of ISAR image frames, as the typical required ISAR aperture angle is 2-3° for the resolutions being considered (see Section \ref{sec:ISAR}). The synthetic aperture size for each ISAR frame is calculated in such a way that the range and cross-range resolutions of the ISAR image are equal, and are constant throughout the sequence. The resulting resolution cells are square and thus analogous to image pixels, allowing image processing techniques which are normally applied to optical imagery to be applied to the ISAR imagery with minimal alteration.

\subsection{ISAR image generation}
\label{sec:ISAR}
The backscattered electric field $E^s$ from a complex target can be closely approximated by the summation of scattering from a large number of single point scatterers, expressed as:
\begin{equation}
    E^s(k, \phi)= \sum_{i=1}^K A_i \cdot \exp \left( -j\,2\, \vec{\mathbf{k}} \cdot \vec{\mathbf{r}}_i \right),
    \label{eq:Es_definition}
\end{equation}
where $k$ is the spatial frequency, $\phi$ is the azimuth angle, $\vec{\mathbf{k}}$ is the spatial wavenumber vector, $\vec{\mathbf{r}}_i~=~x_i\hat{\mathbf{x}}~+~y_i\hat{\mathbf{y}}$ is the position vector of the $i$-th scattering centre with respect to the radar, and $A_i$ is the amplitude of the $i$-th scattering centre.

An ISAR image in the $x$-$y$ image plane is obtained by the 2D inverse Fourier integral of (\ref{eq:Es_definition}) \citep{ozdemir_inverse_2021}: 
\begin{multline}
    \text{ISAR}(x,y) =\iint_{}^{} E^s(k, \phi) \cdot \exp{\left(-j2\,k\,x\right)} \\ \cdot \exp{(-j2k_c\, \phi y)}
    \ dk \ d\phi,
\end{multline}
where $k_c$ is the wavenumber corresponding to centre frequency.
Range resolution $\Delta x$ is defined by signal bandwidth $B$:
\begin{equation}
    \Delta x = \frac{c}{2B},
\end{equation} 
while cross-range resolution $\Delta y$ is defined by centre frequency $f_c$ and integration angle $\Omega$:
\begin{equation}
    \Delta y = \frac{c}{2f_c\Omega}.
\end{equation}

\subsection{Data generation}
\begin{figure*}
    \centering
    \includegraphics[width=1\linewidth]{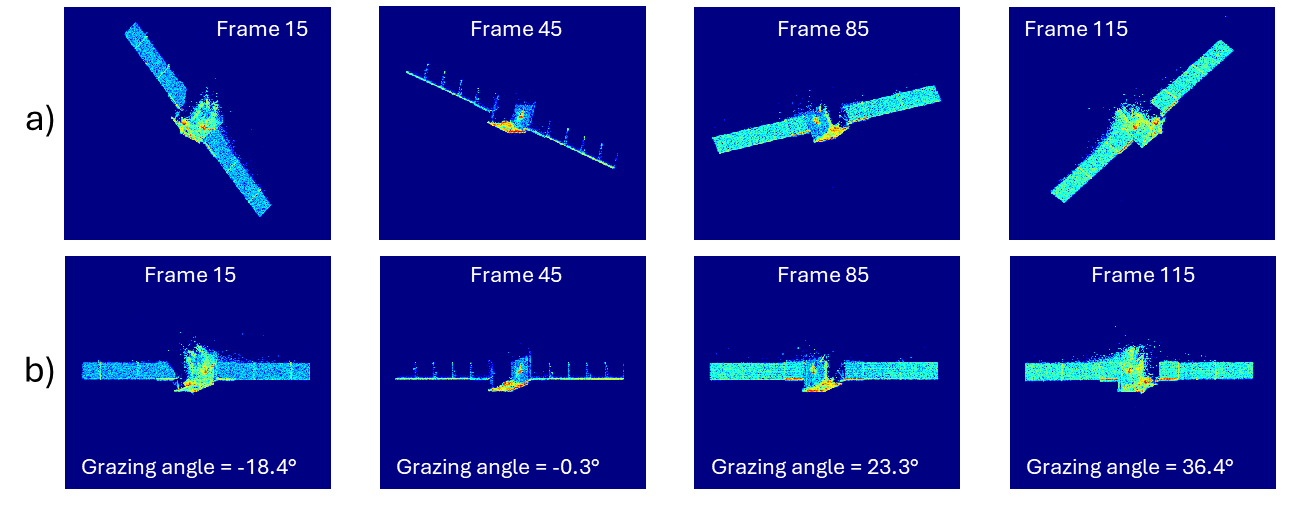}
    \caption{Subset of simulated ISAR images of Skynet 5D at 300 GHz with 5 GHz bandwidth. Each frame has been processed with four-times zero padding and Hanning windowing. Row a) shows unaligned frames, and b) shows their aligned counterparts. The grazing angles relative to the plane of the solar panels is noted.}
    \label{fig:simulation-examples}
\end{figure*}

The data used in this work was generated using the Graphical Electromagnetic ISAR Simulator for sub-THz waves (GEIST), first introduced by \citet{marchetti_electromagnetic_2023} and validated by \citet{jones_novel_2024}. GEIST is able to rapidly simulate high-resolution ISAR imagery, allowing large amounts of data to be produced in a short time. In this work, ISAR images were simulated with GEIST using a bandwidth of 5 GHz, achieving a range resolution of 3 cm. To match cross-range resolution at 300 GHz centre frequency, the integration angle for each frame is 0.95\textdegree. For lower frequencies, the integration angle will be larger.
According to \citet{jones_strategies_2025}, for an average encounter with closest approach distance of 50 km, the total observation angle is $\sim$120\textdegree, resulting in a sequence of $\sim$126 frames covering a significant range of aspect angles. Fig.\ref{fig:simulation-examples} shows a subset of this generated sequence,
annotated with corresponding grazing angles relative to the plane of the solar panels.

In the modelled encounter scenario, the solar panels are oriented parallel to the equatorial plane of the Earth, to provide a useful geometrical reference for the frame sequence generation. On orbit, the spacecraft will be oriented to ensure maximum solar incidence of the panels, which will change the projection of the target spacecraft in the image plane. However, the methods presented in this work are applicable to any mutual attitude of monitoring satellite (MS) and observed satellite (OS).
In the simulation, the solar panel is always in the equatorial plane, but in reality they will move to face the sun.

\subsection{Sequence alignment}
Viewing a target from a range of aspect angles allows its true features to be accentuated in the imagery, and for artefacts and noise to be suppressed. In previous research \citep{jones_strategies_2025}, a method to obtain this constructive information through frame-to-frame alignment within a sequence of ISAR imagery, relying purely on semantic image data, was presented. The method uses a series of affine transformations applied to each image in the original sequence to produce a sequence of aligned images. 
The alignment process is a rule-based expert model, applicable to the majority of existing anthropogenic RSOs.

An affine transformation transforms an original image $I$ into a transformed image $I'$ as:
\begin{equation}
    I'= \textbf{A}I + \Vec{S}
    \label{eq:imagetransform}
\end{equation}
where $ \textbf{A}$ is a matrix comprising rotation, shear, and scale transformations, which can be defined as separate matrices as:
\begin{align}
     \textbf{A} = \underbrace{\begin{pmatrix} c_x & 0 \\ 0 & c_y \end{pmatrix}}_{\text{scale}} 
     \cdot 
     \underbrace{\begin{pmatrix} 1 & d_x \\ d_y & 1 \end{pmatrix}}_{\text{shear}}
     \cdot
     \underbrace{\begin{pmatrix} \cos{\theta} & -\sin{\theta} \\ \sin{\theta} & \cos{\theta} \end{pmatrix}}_{\text{rotation}}
\label{eq:affine}\end{align}

where $c_x$ and $c_y$ are the horizontal and vertical scale factors, $d_x$ and $d_y$ are the horizontal and vertical shear factors, and $\theta$ is the rotation angle. $\Vec{S}$ is the translation vector:
\begin{equation}
    \Vec{S} = \begin{pmatrix} s_x \\s_y \end{pmatrix}
    \label{eq:translation}
\end{equation}
where $s_x$ and $s_y$ are the horizontal and vertical displacements.

For the ISAR sequence alignment, coarse segmentation is initially applied to the image to separate the target from the background, allowing the requisite transformations to be determined and applied after initial attitude axis determination, as in \citet{jones_strategies_2025}.

\section{Feature detection}
\label{sec:feature-detection}
\begin{figure}
    \centering
    \includegraphics[width=1\linewidth]{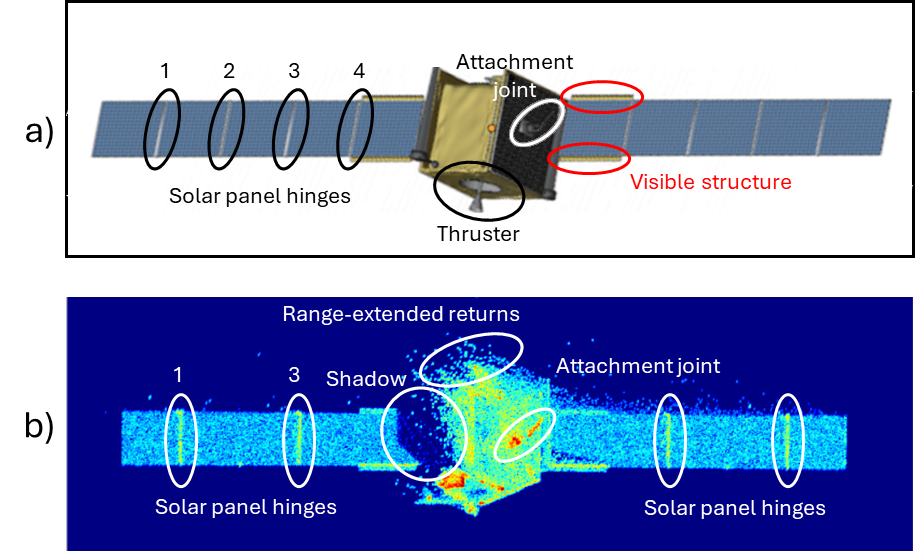}
    \caption{(a) CAD model and (b) ISAR image annotated with examples of potentially identifiable features. In the ISAR image, imaging artefacts (shadowing and range extended returns) are also annotated. The radar imaging plane does not directly correlate to the orientation of the CAD model, which has been chosen to make feature correspondence between CAD and ISAR easily understood.}
    \label{fig:featureCAD-ISAR}
\end{figure}

The purpose of image alignment in this work is to aid correlation of detected features between frames. A feature detected in multiple frames in the same place is likely to be a true feature of the target, while a feature detected in only one frame is more likely a false-positive detection. In Fig. \ref{fig:featureCAD-ISAR}, an example of annotated CAD model (a) and ISAR image (b) are shown to illustrate potentially identifiable features, such as solar panel hinges and satellite thruster. Additionally, imaging artefacts (shadow and range-extended returns) are annotated on the ISAR image. In the CAD model (a), four solar panel hinges can be seen on either side of the body, whereas in the ISAR image (b) only two are visible. On Skynet 5D, the solar panels will have been stowed pre-deployment in a concertina pattern, with hinges protruding on alternating sides. For ISAR, only hinges protruding towards the radar will reflect and appear in the image, hence only alternate hinges are visible in this individual frame. Throughout the sequence, however, different features will become visible as the imaging aspect changes. The combination of this diverse information is discussed later.

Optical edge detection (e.g. Canny's method \citep{canny_computational_1986}) frequently makes use of filter kernels \citep{sobel_gradient_1968, prewitt_object_1970} to calculate image gradients from the difference between neighbouring pixel values, where high gradient values are likely to corresponds to an edge within the image. Gradient calculation is sensitive to noise, so optical images are often smoothed beforehand with a Gaussian filter. \citet{touzi_statistical_1988} showed that when using difference-based operators, the speckle noise in radar imagery results in a greater number of false detections in regions with high reflectivity, rather than a consistent false alarm rate throughout an image. To improve edge detection in radar imagery, \citet{fjortoft_optimal_1998} proposed the use of ratio-based image pixel value gradient calculation, particularly the ratio of exponentially weighted averages (ROEWA).

In this research, the ROEWA method has been applied to the aligned ISAR image to obtain horizontal and vertical image gradients, $G_x$ and $G_y$. These directional gradients are used to calculate total gradient magnitude image $G$ and gradient direction image $\Theta$ as:
\begin{align}
    G=&\sqrt{G_x^2+G_y^2}\\
    \Theta=&\,\text{atan2}(G_x,G_y).
\end{align}

\begin{figure}
    \centering
    \includegraphics[width=1\linewidth]{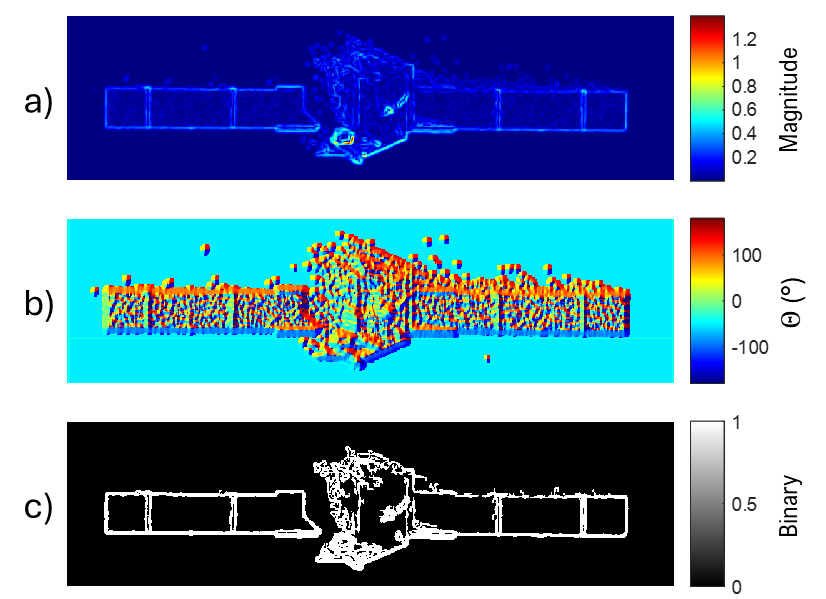}
    \caption{Images showing a) gradient magnitude, b) gradient direction, and c) binary mask of significant regions.}
    \label{fig:magnitudeGradientMask}
\end{figure}

The total gradient magnitude image is thresholded via Otsu's method \citep{otsu_threshold_1975} to obtain a binary mask, and small components are removed to retain a mask of significant regions, which is applied to both the magnitude and orientation images. Fig. \ref{fig:magnitudeGradientMask} shows the original magnitude and gradient images, and the significance mask.
This masking is done to reduce computation time in later processes of linear feature extraction by reducing the number of pixels being considered (in this instance, by a factor of $\sim$30).

As a preliminary example of feature detection, this research focuses on detecting linear features, but the same method is generalisable to other shapes of feature (e.g. elliptical \citep{ballard_generalizing_1981}). To detect linear features within these edge images, a variation of the Hough transform (HT) \citep{hough_method_1962} has been used. This transform is a well-documented computer vision technique, particularly commonplace in linear feature detection \citep{aggarwal_line_2006}, where image space $I(x,y)$ is transformed into Hough space $H(\rho, \theta)$ via parametric representation of a line:
\begin{equation}
    \rho = x \cos{\theta} + y \sin{\theta}.
    \label{eq:Hough-line}
\end{equation}
Points in Hough space with high value are likely to represent sought features in image space, but the transform suffers from interference noise \citep{brown_inherent_1983} meaning that identification of true peaks is challenging. To reduce this noise and improve robustness peak detection, we propose to use a double-weighted Hough transform.

In the standard HT implementation, a binary image's pixels vote evenly in an accumulator to form Hough space. For every pixel $I(x,y)=1$, each corresponding $H(\rho, \theta)$ cell is increased by 1. 
In this implementation, each pixel's vote can instead be weighted by using retained magnitude and direction information. A pixel votes more strongly if it has greater magnitude, and if its direction is aligned with $\theta$. Magnitude weight $W_M$ is the magnitude value ($W_M(x,y) = G(x,y)$), but the interpretation of direction weight $W_D$ is less straightforward. For each pixel, its direction is used to inform a Gaussian distribution $D_G(\theta)$ spanning the range of $\theta$, centred on the pixel's direction. When voting in ($\rho, \theta$) bins, the direction weight $W_D$ is given by the value of $D_G(\theta)$.
These weights are combined into a final vote weight $W=W_M\cdot W_D$.

\begin{figure*}
    \centering
    \includegraphics[width=1\linewidth]{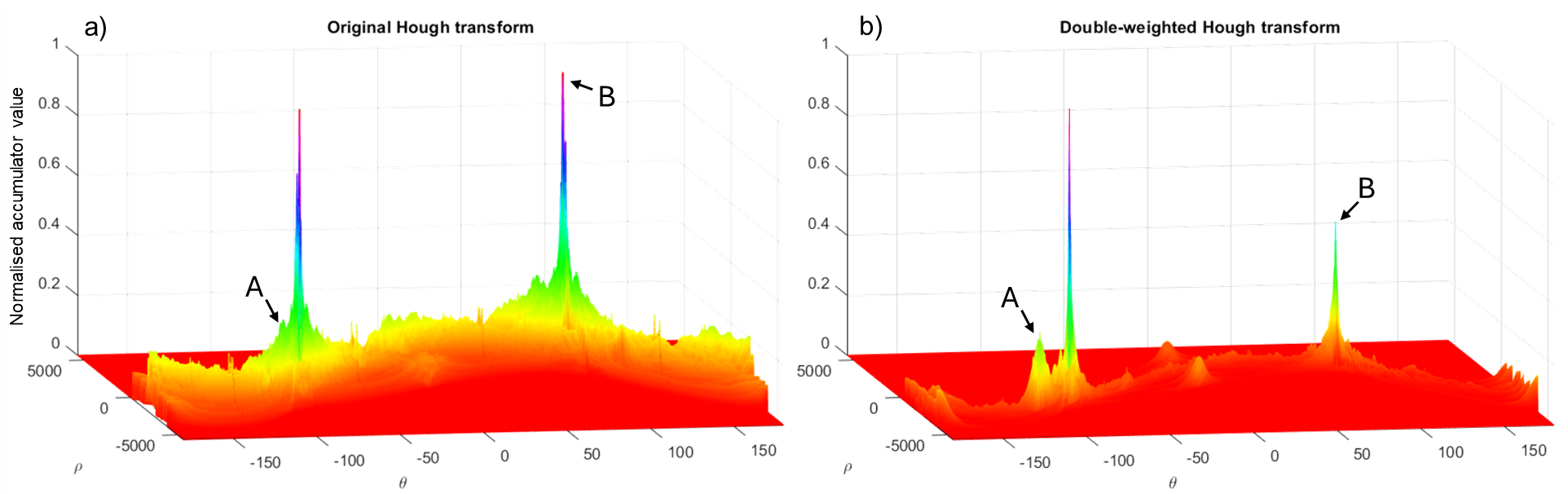}
    \caption{Comparison of (a) standard Hough transform implementation, and (b) the double-weighted Hough transform. In additional to the overall reduction in interference noise, key differences have been annotated: A. a true peak is indistinguishable from noise in (a), but clearly detectable in (b); B. retention of direction information produces unique results across the entire $\theta$ range, rather than duplicated values.}
    \label{fig:houghComparison}
\end{figure*}

Fig. \ref{fig:houghComparison} shows a comparison between the standard HT implementation and the weighted HT, where several major benefits are visible. Noise throughout Hough space  is greatly reduced, making accurate and reliable peak detection significantly easier. Additionally, the retention of direction information renders the entire range $-180$\textdegree\space$<\theta<$ $180$\textdegree\space unique rather than mirroring $0$\textdegree\space $<\theta<$ $180$\textdegree. A detected edge can therefore be identified as rising or falling, which is an important feature descriptor.

With reduced noise, peak detection via local maxima is effective. These peaks are described by ($\rho, \theta$), corresponding to lines of infinite length in image space. To locate a feature, the significance mask (Fig. \ref{fig:magnitudeGradientMask}c) is searched along these lines to find pixels belonging to the feature. These peak locations and corresponding features are used to recognise persistent features.

\section{Persistent feature reconstruction}
\label{sec:persistent}
Detected features with similar properties (e.g. position, orientation) appearing in multiple aligned frames are likely to represent the same true feature of the target, and should be retained. Conversely, a detected feature with no similar detections is likely to be a false detection, and should be discarded. The large quantity of frames obtained from an encounter improves the reliability of persistent feature retention.

This work uses information from the entirety of the sequence for alignment and association, which assumes that an MS-OS encounter is finished before the analysis begins. To identify features and detect anomalies at the earliest opportunity during an MS-OS encounter while on orbit, real-time processing is required. In this case, the alignment process can begin while the encounter is ongoing, with increasing accuracy and reliability of alignment gained by longer sequences. Once the initial set of frames are aligned, information about incoming frames will be accumulated and used in the alignment process.

In Section \ref{sec:feature-clustering}, a method for frame-to-frame feature association is proposed. In Section \ref{sec:sequence-combining}, techniques for combining a sequence into a cumulative image are discussed. In Section \ref{sec:shadow-detection}, a use case for shadow detection via cluster analysis is presented.

\subsection{Frame-to-frame feature association}
\label{sec:feature-clustering}

Features detected by the method described in Section \ref{sec:feature-detection} are parameterised by $\rho$ and $\theta$ (as in Eq. \ref{eq:Hough-line}), and are presented in two-dimensional parameter space as single points. To confirm the features, these points are clustered within this space, based on criteria discussed below. Since the data is unlabelled, and there are an unknown number of true clusters, the unsupervised clustering algorithm ``Density-Based Spatial Clustering of Applications with Noise", or ``DBSCAN" \citep{ester_density-based_1996} has been chosen. DBSCAN provides the additional benefits of being able to classify points belonging to no clusters, and being easily applicable to higher dimensional spaces. This is important because future research will increase the number of descriptive parameters, adding dimensions to parameter space.

\begin{figure}
    \centering
    \includegraphics[width=1\linewidth]{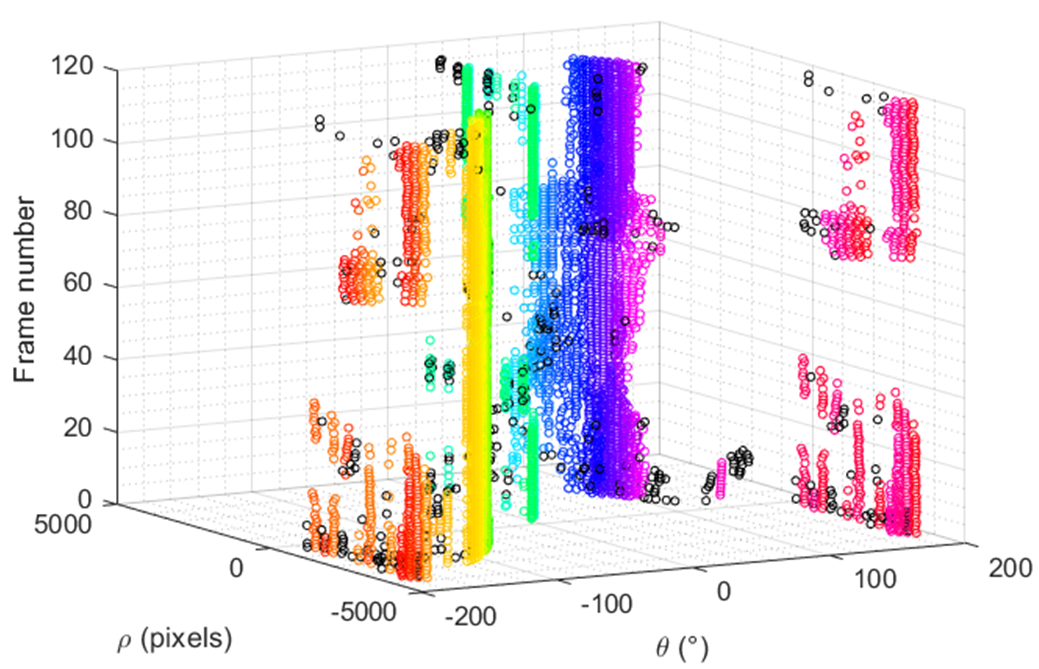} 
    \caption{DBSCAN clustering results, visualised in expanded-3D parameter space. Each point is colour-coded according to its cluster, with unclustered points in black}
    \label{fig:DBSCANclustering}
\end{figure}

The DBSCAN algorithm is controlled by two adjustable parameters, maximum distance between points $\varepsilon$ and minimum number of points per cluster $\mu$, the choices of which heavily affect the accuracy of DBSCAN-based clustering. \citet{sander_density-based_1998} propose setting $\mu=2D$, where $D$ is the number of dimensions, and determining $\varepsilon$ by the elbow point of the \textit{k}-distance plot (the mean distance between each point and its nearest \textit{k} neighbours), where $k=\mu-1$. This approach has been followed. Additionally, since $\varepsilon$ is a distance measurement, the scaling of the axes comprising parameter space is impactful. In this work, $\theta$ and $\rho$ dimensions have been scaled to similar proportions to allocate similar weighting to both during clustering.
The results of DBSCAN clustering can be seen in Fig. \ref{fig:DBSCANclustering}, where the 2D parameter space has been expanded for visualisation by adding a third dimension of the frame number in which the features are detected. To ensure that values either side of $\pm180$\textdegree are clustered together properly, the range of $\theta$ has been increased to nominal $-200$\textdegree\space$<\theta<$ $200$\textdegree, resulting in duplicate values.

For many features, a significant gap appears between frames $\sim$45-75 in the simulated sequence. This is expected, since the MS crosses the plane of the OS's solar panels during the imaging encounter, changing from viewing the satellite structure from the lower aspect to the upper aspect. When the grazing angle from the ISAR sensing system on the MS to the solar panels on the OS is small, the intensities of radar return signals from the majority of the solar panels are too low to appear in imagery, so many features are not detected, resulting in the gap in parameter space. Features that persist throughout the entire sequence include the front edges of the solar panels and the satellite body, appearing in Fig. \ref{fig:DBSCANclustering} as unbroken clusters around $\pm90$\textdegree.
Distinct clusters appear uniquely from the lower aspect (frames 0-45) or the upper aspect (frames 75-120), corresponding to the alternating hinges between solar panel segments, as discussed in Section \ref{sec:feature-detection}.

\begin{figure}
    \centering
    \includegraphics[width=1\linewidth]{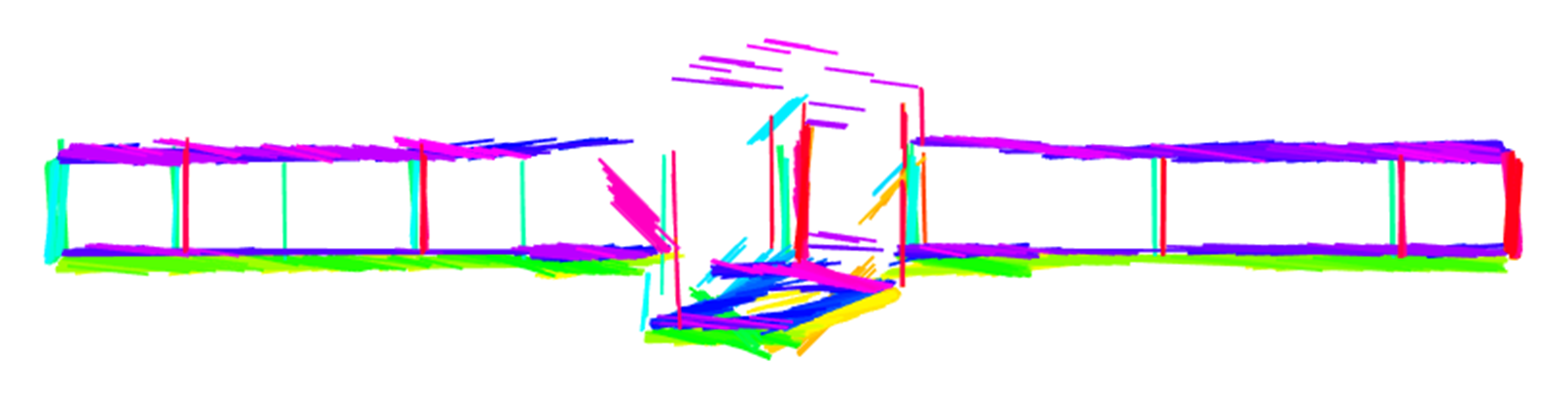}
    \caption{Reconstructed feature map. Each feature is coloured according to its cluster (corresponding with Fig. \ref{fig:DBSCANclustering}).}
    \label{fig:featureMap}
\end{figure}

In Fig. \ref{fig:featureMap}, the detected and clustered features have been reconstructed, forming a map of linear features. When compared to the CAD model in Fig. \ref{fig:featureCAD-ISAR}a, the majority of linear features including the edges of solar panels, hinges between solar panels, edges of satellite body, and solar panel attachment points, have been detected, clustered, and reconstructed successfully. Most of the hinges can be identified clearly in Fig. \ref{fig:featureMap}, despite never being visible simultaneously in the same single image, such as in Fig. \ref{fig:featureCAD-ISAR}b. For four of the six detected hinges the rising and falling edges have been distinguished through the clustering process, shown by the adjacent red (rising) and cyan (falling) feature map markers. This information can be used as a descriptor to classify features in future work.
Each of the features in this map is defined and categorised separately, providing a level of detail about the structure of the target that cannot be extracted from binary edge detection methods alone. 

\subsection{Cumulative image} \label{sec:sequence-combining}
\begin{figure}
    \centering
    \includegraphics[width=1\linewidth]{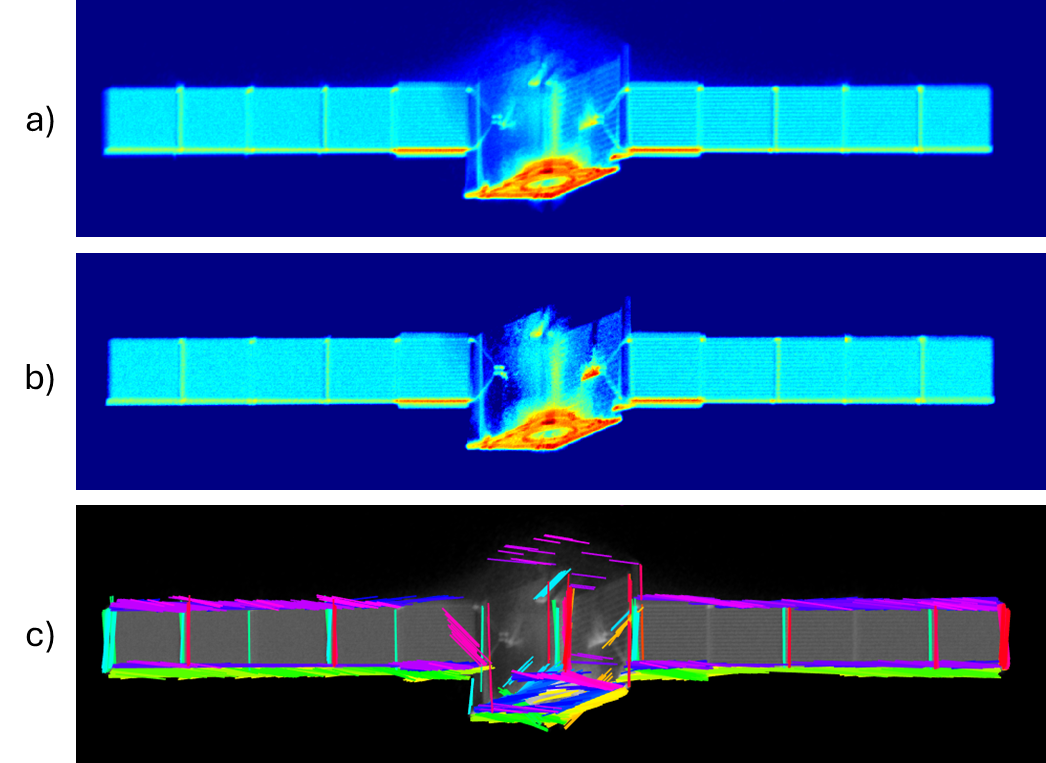}
    \caption{Cumulative image calculated using (a) mean and (b) median. In (c), the mean image has been superimposed with all clustered linear features.}
    \label{fig:compounding-methods}
\end{figure}

Throughout the sequence, image artefacts such as range extended returns and sidelobes will vary from frame to frame. Combining the frames into a single image will reduce the impact of these artefacts. Since these artefacts are aligned with the radar boresight direction, the rotation of each frame by a unique angle during the alignment process decorrelates the artefacts by altering the imaging planes.
The intensity of a pixel in the cumulative image is calculated by use of either the mean or median values of that pixel across all frames. The resulting mean and median cumulative images are shown in Fig. \ref{fig:compounding-methods}, along with a superimposition of the feature map (Fig. \ref{fig:featureMap}) onto the mean image.
In the mean-valued image, the noise behind the satellite body is present, while it has been completely removed from the median-valued image. Both images reveal features that would be difficult to identify in a single frame: the attachments between the solar panels and the body, the extent of the thruster cavity, and regular narrow strips along the solar panels. Although this cavity is not explicitly highlighted by linear features, in future work, elliptical feature detection will be used to highlight it.

\subsection{Shadow detection use case}
\label{sec:shadow-detection}
To demonstrate the efficacy of the proposed approach, we will consider a special class of linear feature representing an edge of a shadow cast upon the satellite via self-occlusion.
Shadows in ISAR imagery are challenging to separate confidently from background noise, either by statistical or semantic classification.
However, a shadow edge in an aligned ISAR image sequence can be distinguished from an actual physical feature by exhibiting both of the following attributes:
\begin{enumerate}
    \item A high-contrast boundary between the shadowed region and neighbouring illuminated regions, detected as an edge with a high gradient, accurately localised within an image.
    \item Sequence alignment results in physical features remaining static throughout the sequence, but shadows will appear to move \citep{kumar_imaging_2025}. This distinguishes shadows from physical features exhibiting the previous attribute, such as regions of damage and external target edges.
\end{enumerate}

\begin{figure}
    \centering
    \includegraphics[width=1\linewidth]{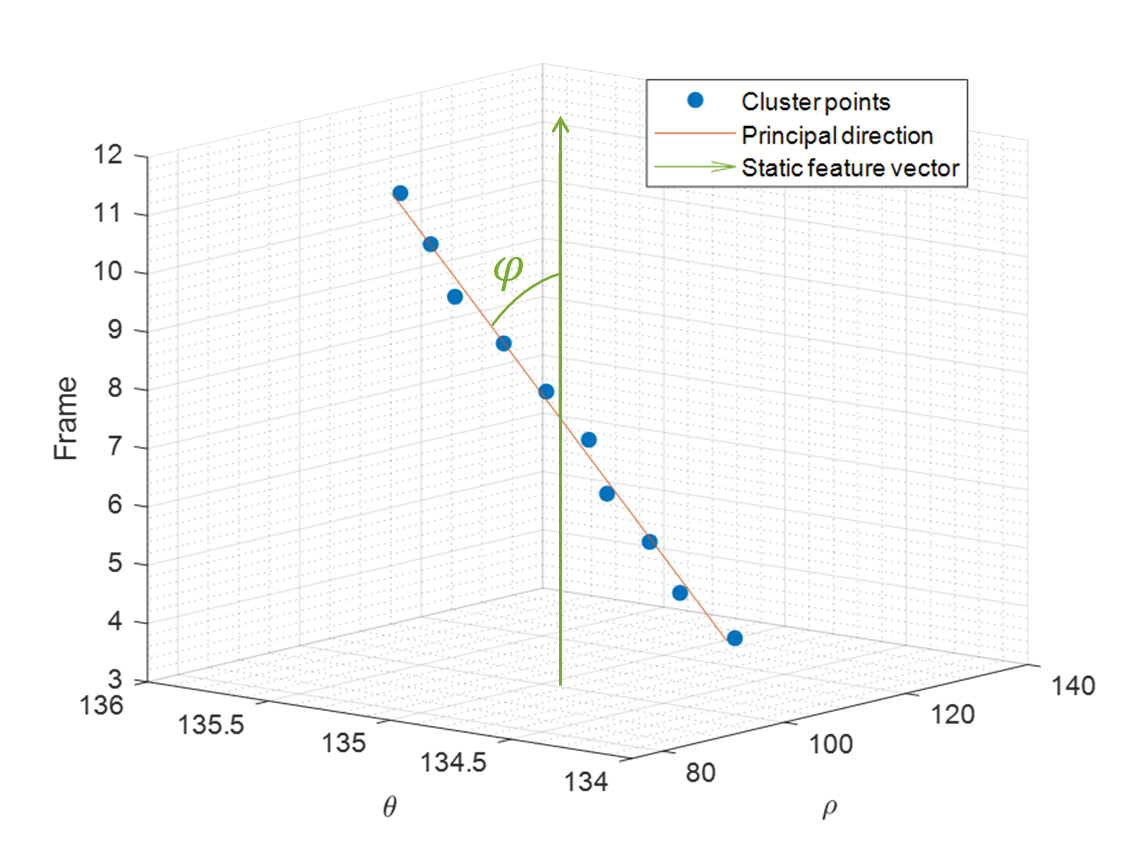}
    \caption{Example of 3D parameter space for an individual cluster. PCA is used to calculate the principal direction vector (orange). The divergence angle $\varphi$ is measured between this vector and the ideal static feature vector (green).}
    \label{fig:PCAcluster}
\end{figure}

To devise a descriptor of shadow edge based on both these attributes, the evolution of detected features throughout the sequence must be considered, which can be measured as the change in $\rho$ and $\theta$ with respect to frame number. Reconstructing the 3D parameter space (seen in Fig. \ref{fig:DBSCANclustering}) for individual clusters, principal component analysis (PCA, \citet{abdi_principal_2010}) can be used to calculate a cluster's principal direction vector and the proportion of variance $P_V$ explained by this direction, which is a measure of how much of the total data variability this direction captures.
An ideal static feature will exhibit no change in $\rho$ and $\theta$, resulting in a vector normal to the $\rho$-$\theta$ plane. The divergence angle, $\varphi$, between the cluster vector and an ideal static feature's vector is a measure of the feature's rate of motion throughout the sequence. Fig. \ref{fig:PCAcluster} shows the parameter space for an example cluster, with the principal direction vector and ideal static feature vector illustrated.
In practical applications, a static feature will not match the ideal case, and will have some small $\varphi$. A shadow's $\varphi$ will be much greater, so a minimum threshold for consideration can be implemented.

For a feature's evolution to be analysed with confidence, it should appear in a minimum number of frames throughout the sequence. Additionally, since shadow edges are well-defined by high constrast, $P_V$ should be close to 1.
Threshold values for the number of frames in which a feature appears and $P_V$ can be adjusted for specific use-cases. In the presented case, based on a trial approach, the minimum number of frame appearances for confident shadow edge tracking was chosen to be 10.

\begin{figure}
    \centering
    \includegraphics[width=1\linewidth]{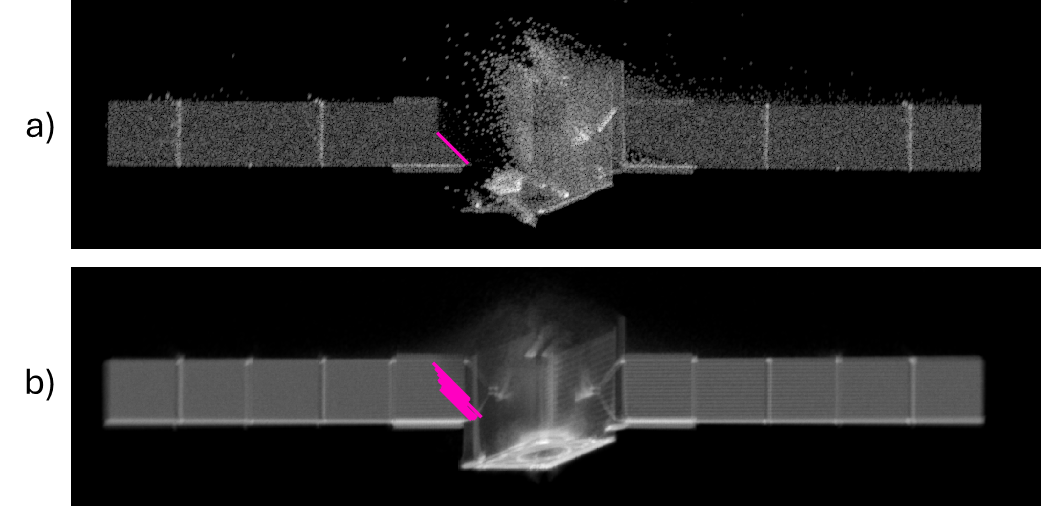}
    \caption{Detected shadow feature illustration: a) single feature superimposed on corresponding image frame, and b) feature evolution superimposed on sequence mean image.}
    \label{fig:specificCluster}
\end{figure}

The cluster with the highest $P_V$ ($=0.999$) and a significant $\varphi$ ($=80.7$\textdegree) has been chosen for illustration. This is the same cluster as shown in Fig. \ref{fig:PCAcluster}.
An overlay of the detected feature corresponding to this cluster is shown in Fig. \ref{fig:specificCluster}. By comparing Fig. \ref{fig:specificCluster}a with Fig. \ref{fig:featureCAD-ISAR}b, it can be seen that this feature demarcates the edge of a shadow visible in certain frames of the sequence. The evolution of this feature is shown in Fig. \ref{fig:specificCluster}b. 

If a feature is moving throughout the aligned sequence, it is neither damage nor structure and can be identified as shadow.
Subsequently, the identification of the shadow can be used to robustly determine information about the occluding object or objects, including association with specific spacecraft structures, such as the bus or deployable elements.

\section{Conclusion}
This work has presented a novel approach using several methods for extracting information about the external infrastructure of RSOs from sequences of high-resolution ISAR imagery. 
These methods, including image alignment, Hough Transform, DBSCAN, and feature descriptor identification provide a robust tool for revealing features that are obscured or otherwise undetectable in individual frames, enabling comprehensive information to be obtained about the entirety of the target's structure and improving confidence in feature detection and tracking. This knowledge will inform characterisation and classification of RSO condition and functionality, a key capability for improving space domain awareness.

Future directions of study include enhancement and expansion of feature detection, use of higher order models for linear feature identification such as the generalised Hough Transform, extended Kalman Filter utility for enhanced feature tracking and characterisation, and incorporation of feature detection with previous segmentation and classification work \citep{coe_segmentation_2024}. These attributes will underpin the development of a multi-modal ISAR image classifier.
\section*{Acknowledgments}
This work is supported by the Engineering and Physical Sciences Research Council (EPSRC) grant (EP/Y022092/1) and the Midlands Space Cluster. It is also supported by an EPSRC DTP scholarship, and a Defence Science and Technology laboratory (DSTL) scholarship ``THzISAR - Terahertz ISAR image formation for Space-to-Space (Sat$^2$) Intelligence, Surveillance, and Recognisance (ISR)"

\bibliographystyle{jasr-model5-names}
\biboptions{authoryear}
\bibliography{references2}

@misc{european_space_agency_esa_2025,
	title = {{ESA} {Space} {Environment} {Report} 2025},
	language = {en},
	urldate = {2025-07-02},
	author = {ESA},
	year = {2025},
}

@article{aggarwal_line_2006,
	title = {Line detection in images through regularized hough transform},
	volume = {15},
	issn = {1941-0042},
	number = {3},
	urldate = {2025-06-02},
	journal = {IEEE Trans. Image Process.},
	author = {Aggarwal, N. and Karl, W.C.},
	month = mar,
	year = {2006},
	pages = {582--591},
}

@article{brown_inherent_1983,
	title = {Inherent {Bias} and {Noise} in the {Hough} {Transform}},
	volume = {PAMI-5},
	issn = {1939-3539},
	number = {5},
	urldate = {2025-06-02},
	journal = {IEEE Trans. Pattern Anal. Mach. Intell.},
	author = {Brown, Christopher M.},
	month = sep,
	year = {1983},
	pages = {493--505},
}

@book{chen_inverse_2014,
	title = {Inverse {Synthetic} {Aperture} {Radar} {Imaging}},
	isbn = {978-1-61353-013-9 978-1-61353-035-1},
	language = {en},
	urldate = {2025-08-06},
	publisher = {Institution of Engineering and Technology},
	author = {Chen, Victor C. and Martorella, Marco},
	month = sep,
	year = {2014},
}

@inproceedings{coe_segmentation_2024,
	title = {Segmentation and {Classification} of {Sub}-{THz} {ISAR} {Imagery}},
	urldate = {2025-07-03},
	booktitle = {2024 {International} {Radar} {Symposium} ({IRS})},
	author = {Coe, Morgan and Jones, Gruffudd and Gashinova, Marina and Cherniakov, Mikhail and Martorella, Marco and Alconcel, Leah-Nani and Marchetti, Emidio},
	month = jul,
	year = {2024},
	pages = {233--238},
}

@inproceedings{du_swin_2024,
	address = {Zhuhai, China},
	title = {Swin {Transformer}-{Based} {Unsupervised} {Domain} {Adaptation} for {ISAR} {Satellite} {Classification}},
	copyright = {https://doi.org/10.15223/policy-029},
	isbn = {9798331515669},
	language = {en},
	urldate = {2025-05-12},
	booktitle = {2024 {IEEE} {International} {Conf.} {Signal}, {Information} and {Data} {Processing} ({ICSIDP})},
	publisher = {IEEE},
	author = {Du, Liang and Cheng, Qiang},
	month = nov,
	year = {2024},
	pages = {1--6},
}

@article{sander_density-based_1998,
	title = {Density-{Based} {Clustering} in {Spatial} {Databases}: {The} {Algorithm} {GDBSCAN} and {Its} {Applications}},
	volume = {2},
	issn = {1573-756X},
	shorttitle = {Density-{Based} {Clustering} in {Spatial} {Databases}},
	language = {en},
	number = {2},
	urldate = {2025-04-02},
	journal = {Data Mining and Knowledge Discovery},
	author = {Sander, Jörg and Ester, Martin and Kriegel, Hans-Peter and Xu, Xiaowei},
	month = jun,
	year = {1998},
	pages = {169--194},
}

@article{fjortoft_optimal_1998,
	title = {An optimal multiedge detector for {SAR} image segmentation},
	volume = {36},
	issn = {1558-0644},
	number = {3},
	urldate = {2025-09-25},
	journal = {IEEE Trans. Geosci. Remote Sens.},
	author = {Fjortoft, R. and Lopes, A. and Marthon, P. and Cubero-Castan, E.},
	month = may,
	year = {1998},
	pages = {793--802},
}

@patent{hough_method_1962,
	title = {Method and means for recognizing complex patterns},
	nationality = {US},
	assignee = {Individual},
	number = {US3069654A},
	urldate = {2025-06-02},
	author = {Hough, Paul V. C.},
	month = dec,
	year = {1962},
}

@article{jones_strategies_2025,
	title = {Strategies for {Monitoring} of {Assets} in {Geosynchronous} {Orbit} ({GEO}) {Using} {Space}-{Based} {Sub}-{THz} {Inverse} {Synthetic} {Aperture} {Radar} ({ISAR})},
	volume = {3},
	issn = {2832-7357},
	urldate = {2025-06-23},
	journal = {IEEE Trans. Radar Syst.},
	author = {Jones, Gruffudd and Coe, Morgan and Beesley, Lily and Alconcel, Leah-Nani and Martorella, Marco and Gashinova, Marina},
	year = {2025},
	pages = {656--667},
}

@inproceedings{jones_novel_2024,
	title = {Novel {Simulation} {Method} for {Sub}-{THz} {ISAR} {Imaging} of {Space} {Objects}},
	urldate = {2025-02-20},
	booktitle = {2024 21st {European} {Radar} {Conference} ({EuRAD})},
	author = {Jones, Gruffudd and Coe, Morgan and Marchetti, Emidio and Alconcel, Leah-Nani and Cherniakov, Mikhail and Gashinova, Marina},
	month = sep,
	year = {2024},
	pages = {272--275},
}

@article{ballard_generalizing_1981,
	title = {Generalizing the {Hough} transform to detect arbitrary shapes},
	volume = {13},
	issn = {0031-3203},
	number = {2},
	urldate = {2025-12-05},
	journal = {Pattern Recognition},
	author = {Ballard, D. H.},
	month = jan,
	year = {1981},
	pages = {111--122},
}

@article{klare_future_2024,
	title = {The {Future} of {Radar} {Space} {Observation} in {Europe}—{Major} {Upgrade} of the {Tracking} and {Imaging} {Radar} ({TIRA})},
	volume = {16},
	copyright = {http://creativecommons.org/licenses/by/3.0/},
	issn = {2072-4292},
	language = {en},
	number = {22},
	urldate = {2025-08-05},
	journal = {Remote Sensing},
	author = {Klare, Jens and Behner, Florian and Carloni, Claudio and Cerutti-Maori, Delphine and Fuhrmann, Lars and Hoppenau, Clemens and Karamanavis, Vassilis and Laubach, Marcel and Marek, Alexander and Perkuhn, Robert and Reuter, Simon and Rosebrock, Felix},
	month = jan,
	year = {2024},
	pages = {4197},
}

@article{li_polarimetric_2025,
	title = {Polarimetric {ISAR} {Space} {Target} {Structure} {Recognition} {Based} on {Embedded} {Scattering} {Mechanism} and {Semi}-{Supervised} {Representation} {Learning}},
	volume = {63},
	copyright = {https://ieeexplore.ieee.org/Xplorehelp/downloads/license-information/IEEE.html},
	issn = {0196-2892, 1558-0644},
	language = {en},
	urldate = {2025-05-12},
	journal = {IEEE Trans. Geosci. Remote Sens.},
	author = {Li, Ming-Dian and Xiao, Shun-Ping and Chen, Si-Wei},
	year = {2025},
	pages = {1--19},
}

@misc{macdonald_husir_2014,
	title = {The {HUSIR} {W}-{Band} {Transmitter}},
	language = {en},
	author = {MacDonald, Michael E and Anderson, James P and Lee, Roy K and Gordon, David A and McGrew, G Neal},
	year = {2014},
}

@inproceedings{marchetti_electromagnetic_2023,
	title = {Electromagnetic simulator based on graphical computing and physical optics for sub-{THz} {ISAR} imagery of space objects},
	urldate = {2025-10-21},
	booktitle = {2023 24th {International} {Radar} {Symposium} ({IRS})},
	author = {Marchetti, E. and Hoare, E. and Cherniakov, M. and Gashinova, M.},
	month = may,
	year = {2023},
	pages = {1--9},
}

@article{marchetti_space-based_2022,
	title = {Space-{Based} {Sub}-{THz} {ISAR} for {Space} {Situational} {Awareness} - {Laboratory} {Validation}},
	volume = {58},
	issn = {1557-9603},
	number = {5},
	urldate = {2024-04-17},
	journal = {IEEE Trans. Aerosp. Electron. Syst.},
	author = {Marchetti, Emidio and Stove, Andrew G. and Hoare, Edward G. and Cherniakov, Mikhail and Blacknell, David and Gashinova, Marina},
	month = oct,
	year = {2022},
	pages = {4409--4422},
}

@article{otsu_threshold_1975,
	title = {A {Threshold} {Selection} {Method} from {Gray}-{Level} {Histograms}},
	volume = {11},
	language = {en},
	number = {285-296},
	journal = {Automatica},
	author = {Otsu, Nobuyuki},
	year = {1975},
	pages = {23--27},
}

@article{touzi_statistical_1988,
	title = {A statistical and geometrical edge detector for {SAR} images},
	volume = {26},
	issn = {1558-0644},
	number = {6},
	urldate = {2025-10-16},
	journal = {IEEE Trans. Geosci. Remote Sens.},
	author = {Touzi, R. and Lopes, A. and Bousquet, P.},
	month = nov,
	year = {1988},
	pages = {764--773},
}

@article{xin_scale-shift_2025,
	title = {Scale-{Shift} {Attention} in {Polarization} {Domain} for {Fine}-{Grained} {Classification} of {Satellite} {ISAR} {Images}},
	copyright = {https://ieeexplore.ieee.org/Xplorehelp/downloads/license-information/IEEE.html},
	issn = {1520-9210, 1941-0077},
	language = {en},
	urldate = {2025-05-12},
	journal = {IEEE Trans. Multimedia},
	author = {Xin, Zewei and Li, Qinya and Sheng, Bowen and Wu, Fan and Chen, Guihai},
	year = {2025},
	pages = {1--10},
}

@article{xue_sequential_2022,
	title = {Sequential {ISAR} {Target} {Classification} {Based} on {Hybrid} {Transformer}},
	volume = {60},
	issn = {1558-0644},
	urldate = {2025-08-04},
	journal = {IEEE Trans. Geosci. Remote Sens.},
	author = {Xue, Ruihang and Bai, Xueru and Cao, Xiangyong and Zhou, Feng},
	year = {2022},
	pages = {1--11},
}

@inproceedings{kumar_imaging_2025,
	title = {Imaging {Shadows} with {Sub}-{THz} {ISAR} for {Space}-{Borne} {Targets}},
	urldate = {2025-12-08},
	booktitle = {2025 26th {International} {Radar} {Symposium} ({IRS})},
	author = {Kumar, Dillon and Jones, Gruffudd and Coe, Morgan and Beesley, Lily and Alconcel, Leah-Nani and Cherniakov, Mikhail and Gashinova, Marina},
	month = may,
	year = {2025},
	pages = {1--6},
}

@article{abdi_principal_2010,
	title = {Principal component analysis},
	volume = {2},
	copyright = {Copyright © 2010 John Wiley \& Sons, Inc.},
	issn = {1939-0068},
	language = {en},
	number = {4},
	urldate = {2025-12-08},
	journal = {WIREs Computational Statistics},
	author = {Abdi, Hervé and Williams, Lynne J.},
	year = {2010},
	pages = {433--459},
}

@book{ozdemir_inverse_2021,
	title = {Inverse {Synthetic} {Aperture} {Radar} {Imaging} {With} {MATLAB} {Algorithms}},
	isbn = {978-1-119-52133-4},
	language = {en},
	publisher = {John Wiley \& Sons},
	author = {Ozdemir, Caner},
	month = may,
	year = {2021},
}

@inproceedings{ester_density-based_1996,
	title = {A density-based algorithm for discovering clusters in large spatial databases with noise},
	volume = {96},
	booktitle = {kdd},
	author = {Ester, Martin and Kriegel, Hans-Peter and Sander, Jörg and Xu, Xiaowei and {others}},
	year = {1996},
	pages = {226--231},
}

@article{canny_computational_1986,
	title = {A {Computational} {Approach} to {Edge} {Detection}},
	volume = {PAMI-8},
	issn = {1939-3539},
	number = {6},
	urldate = {2025-12-15},
	journal = {IEEE Trans. Pattern Anal. Mach. Intell.},
	author = {Canny, John},
	month = nov,
	year = {1986},
	pages = {679--698},
}

@article{sobel_gradient_1968,
  title={A 3x3 isotropic gradient operator for image processing},
  author={Sobel, Irwin and Feldman, Gary and others},
  journal={a talk at the Stanford Artificial Project in},
  volume={1968},
  pages={271--272},
  year={1968}
}

@article{prewitt_object_1970,
	title = {Object enhancement and extraction},
	volume = {10},
	number = {1},
	journal = {Picture processing and Psychopictorics},
	author = {Prewitt, Judith MS and others},
	year = {1970},
	pages = {15--19},
}
\newpage

\end{document}